%% file: root.tex
\documentclass[letterpaper, 10 pt, conference]{ieeeconf} 
\IEEEoverridecommandlockouts              
\usepackage{graphics}
\usepackage{epsfig}
\usepackage{mathptmx}
\usepackage{times}
\usepackage{amsmath}
\usepackage{amssymb}
\usepackage{tabularx}
\usepackage{multicol}
\usepackage[bookmarks=true]{hyperref}
\usepackage{graphicx}
\usepackage{float}
\usepackage{wrapfig}
\usepackage{lipsum}
\usepackage{float}
\restylefloat{figure}
\graphicspath{{./im/}}
\usepackage[font=small]{caption}
\usepackage[font=small]{subcaption}
\usepackage[draft]{pgf}
\usepackage{mathtools}
\usepackage{amsmath}
\usepackage{hyperref}
\usepackage{color}
\usepackage{sidecap}
\usepackage{subcaption}
\usepackage{colortbl}

\title{\LARGE \bf
Robot Navigation in Constrained Pedestrian Environments using Reinforcement Learning
}

\author{
    Claudia P\'{e}rez-D'Arpino,  
	Can Liu,
	Patrick Goebel,
	Roberto Mart\'{i}n-Mart\'{i}n,
	Silvio Savarese
\thanks{Authors are with the Department of Computer Science at Stanford University.}
}

\begin{document}
\belowdisplayskip=4pt plus 3pt minus 4pt
\belowdisplayshortskip=2pt plus 3pt minus 4pt

\maketitle
\thispagestyle{empty}
\pagestyle{empty}

\begin{abstract} 
    Navigating fluently around pedestrians is a necessary capability for mobile robots deployed in human environments, such as buildings and homes. While research on social navigation has focused mainly on the scalability with the number of pedestrians in \textit{open spaces}, typical indoor environments present the additional challenge of \textit{constrained spaces} such as corridors and doorways that limit maneuverability and influence patterns of pedestrian interaction. We present an approach based on reinforcement learning (RL) to learn policies capable of dynamic adaptation to the presence of moving pedestrians while navigating between desired locations in constrained environments. The policy network receives guidance from a motion planner that provides waypoints to follow a globally planned trajectory, whereas RL handles the local interactions. We explore a compositional principle for multi-layout training and find that policies trained in a small set of geometrically simple layouts successfully generalize to more complex unseen layouts that exhibit composition of the structural elements available during training. Going beyond walls-world like domains, we show transfer of the learned policy to unseen 3D reconstructions of two real environments. These results support the applicability of the compositional principle to navigation in real-world buildings and indicate promising usage of multi-agent simulation within reconstructed environments for tasks that involve interaction. \\ 
    {\footnotesize \href{https://ai.stanford.edu/~cdarpino/socialnavconstrained/}{{\color{blue} https://ai.stanford.edu/$\sim$cdarpino/socialnavconstrained/}}}
\end{abstract}

\section{Introduction}

\begin{figure}[t]
\centering
\includegraphics[width=1\linewidth]{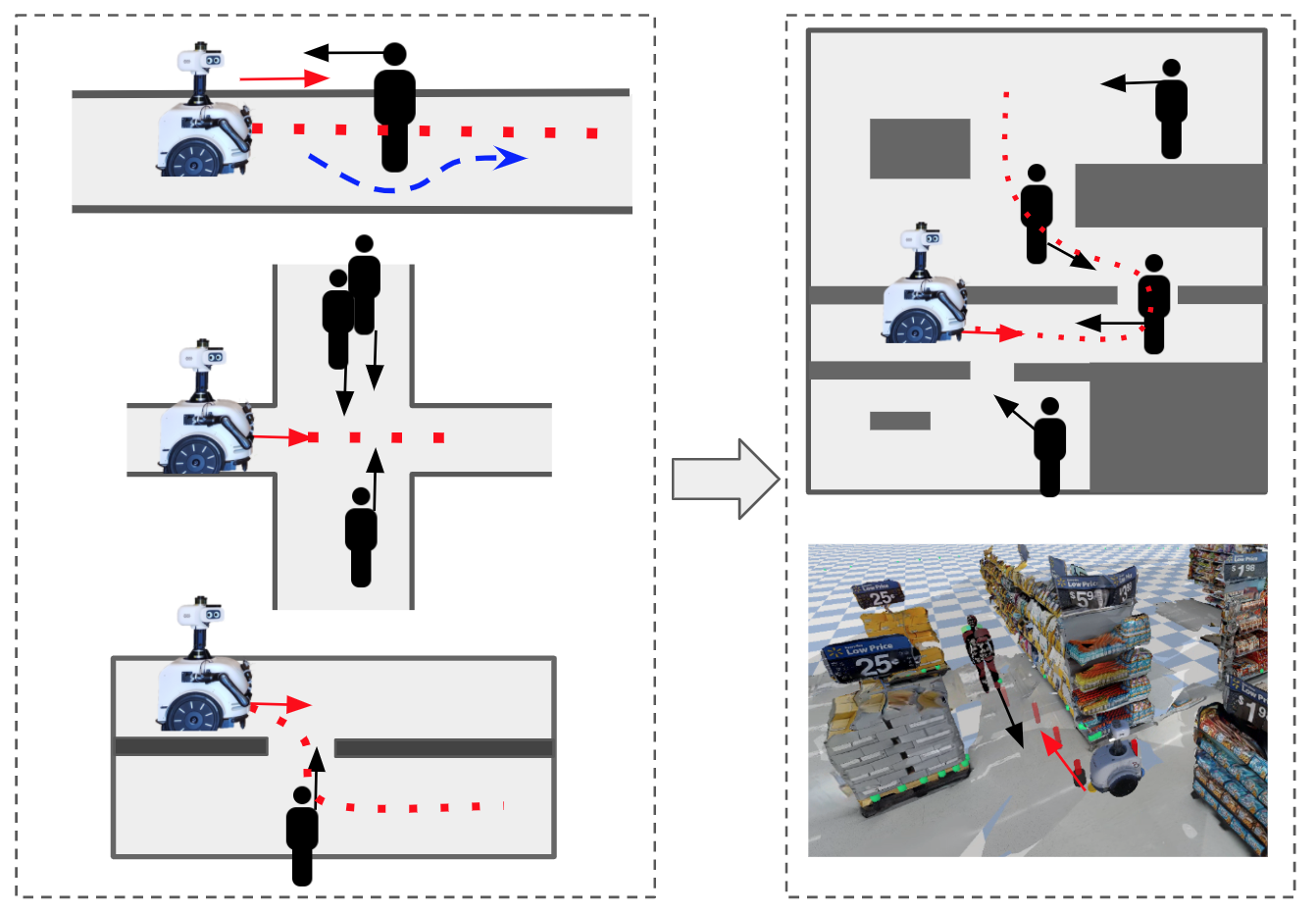}
\caption{
Illustration of the problem of robot navigation around pedestrians in constrained indoor environments. The robot follows a global path (red dotted line) while maneuvering to circulate seamlessly around pedestrians (blue dashed line example). 
The robot's policy is trained on a small set of simple walls-world environments representative of common indoor layouts and the emerging interactions, such as in corridors and intersections (left). We show generalization to two types of unseen environments (right): more complex compositional walls-worlds (top right) and 3D reconstructions of real-world sites (bottom right).
\vspace{-5mm}
}
\label{fig:intro3}
\end{figure}

As mobile robots become more capable of executing a variety of tasks, they are becoming increasingly likely to be deployed in environments co-located with people. This step demands seamless and safe integration into spaces with pedestrians such as buildings, homes and offices. In the path towards socially compliant mobile robots, research on human-robot co-navigation 
has made progress on planning and learning approaches 
for robots that can handle local interactions with moving pedestrians among crowds in open spaces.
Navigation in indoor environments presents additional challenges due to space constraints that limit maneuverability around the pedestrians and create layout-dependent patterns of pedestrian interaction.
These spaces, such as corridors, doors and intersections pose  hard problems for any existing navigation solution. For these types of spaces, there is a need for new interactive navigation models for tight spaces with constraints imposed by obstacles, walls, doors and other common structural elements, based only on on-board sensor information. Fig. \ref{fig:intro3} illustrates some of the basic layouts and the common interactions that emerge in these geometries from typical indoor environments. 

In this paper we present a novel robot learning approach and training strategy focusing in the ability to navigate around pedestrians in typical indoor environments 
in which spatial constraints are sufficiently important to influence the feasibility of the robot's behavior. 
The proposed architecture takes advantage of the complementary strengths of
motion planning and reinforcement learning (RL),
in which the RL 
component learns to handle the local interactions with pedestrians as it pursues the globally planned trajectory, and
adapts to the current conditions of the environment 
as observed with on-board sensors, even in constrained indoor spaces such as corridors and doors.
We observe that typical indoor environments exhibit common geometric elements and diverse composition of these elements. With the aim of generalization to novel realistic spaces, we explore a compositional principle for multi-layout training, in which a policy is trained in a small number of simple, canonical, layouts, such as a corridor, or a single door exit (Fig. \ref{fig:intro3} left). We analyze the capability of this approach to generalize to more complex layouts with multiple composed and combined navigation challenges not seen during training (Fig. \ref{fig:intro3} top right), and find experimentally that policies trained on a set simple layouts generalize better when those represent the elements of the geometries and interactions that conform to the target layout.
This insight results particularly relevant for the   built environment, which is typically composed by combinations of basic layouts templates, such as hallways, rooms, door exits, and crosswalks. 
We then demonstrate transfer of the policies learned on walls-worlds (Fig. \ref{fig:intro3} left) to unseen 3D reconstructions of two real-world environments (Fig. \ref{fig:intro3} bottom right), such as a supermarket and an apartment.

\begin{figure}[t]
  \centering        
  \includegraphics[width=0.48\textwidth]{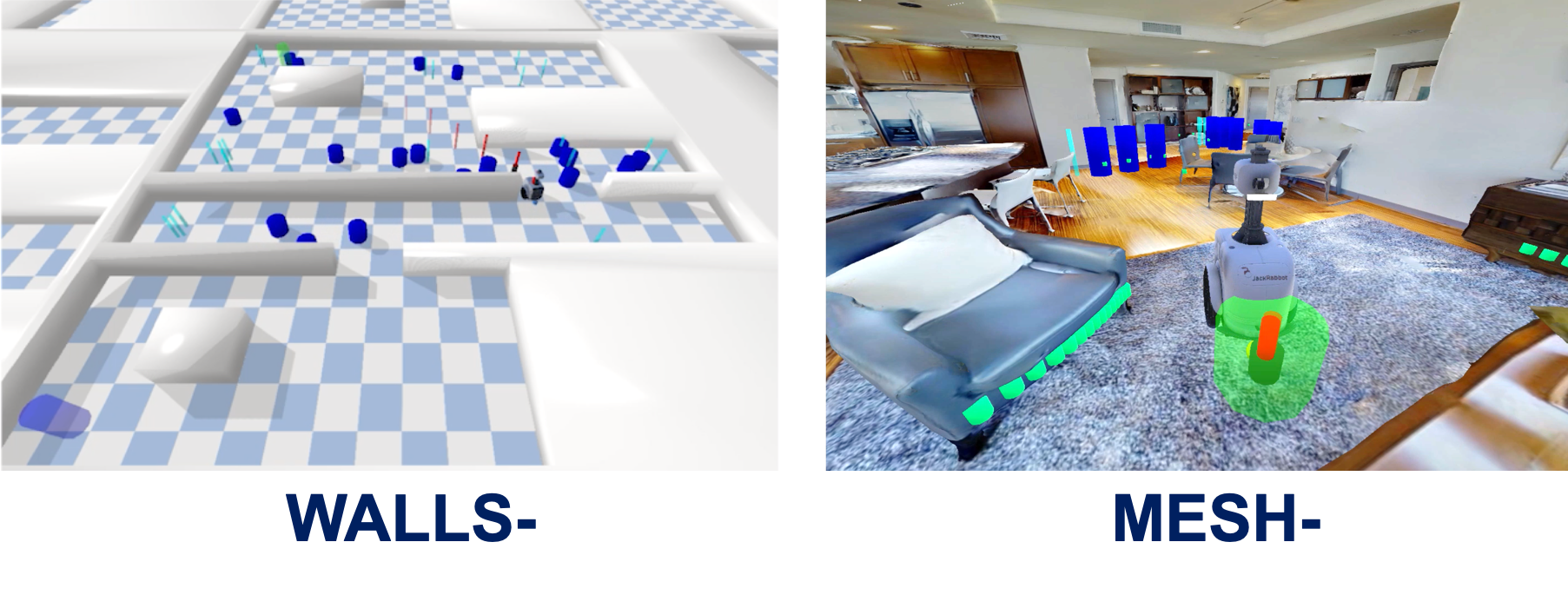}
  \caption{Implemented multi-agent simulation environments. Layouts with the prefix \textbf{WALLS-} are simple walls-worlds (left). The prefix \textbf{MESH-} is used for 3D reconstructions from real sites (right). \vspace{-7mm}}
  \label{fig:wallsmesh}
\end{figure}

We instantiate our navigation approach and train it to control a simulated non-holonomic wheeled robot in multiple environments on the interactive Gibson simulator \cite{xiazamirhe2018gibsonenvCVPR}\cite{48697interactivegibsonRAL}, which leverages the Pybullet physics engine~\cite{coumans2016pybullet} for realistic physics simulation and supports virtual sensing (lidar) for our multi-agent setup. 
The environment includes  simulated pedestrians driven by the Optimal Reciprocal Collision Avoidance (ORCA) model ~\cite{van2011reciprocalORCA} based on a social forces model~\cite{helbing1995socialForcesModel}.
Fig. \ref{fig:wallsmesh} illustrates the two types of simulation environments used in this paper:
the simple walls-worlds layouts implemented using simple shapes, denominated by the prefix WALLS- (Fig. \ref{fig:wallsmesh} left), and the real-scale 3D reconstructed environments implemented using textured meshes from the Gibson dataset \cite{xiazamirhe2018gibsonenvCVPR} (Fig. \ref{fig:wallsmesh} right), denominated by the prefix MESH-.

In summary, this paper presents the following contributions: 
(1) a \textbf{learning approach and a simulation environment} 
for the problem of goal-oriented robot navigation around moving pedestrians,
that unifies ideas from reinforcement learning, planning and multi-agent simulation
to enable human-robot co-navigation 
in constrained environments representative of real human indoor spaces;
(2) \textbf{a compositional multi-layout training regime} using canonical walls-worlds layouts that generalizes to more complex environments; 
and (3) a \textbf{demonstration of transfer} of the learned policy on simplified walls-world \textbf{to unseen 3D reconstructions of two real-world environments} represented by scanned 3D textured meshes. These results support the compositional principle applicability to real-world environments and indicate promising frontiers for the emerging topic of agent simulation within reconstructed environments for tasks that involve human-robot interaction. Accompanying the paper, we present a \href{https://ai.stanford.edu/~cdarpino/socialnavconstrained/}{video} \cite{PaperURLandVideo2020} with demonstrations of the robot behaviors obtained in the experiments.

\section{Related Work}
\label{sec:relatedwork}

\textbf{\textit{Navigating safely among humans}} is a prerequisite for the deployment of mobile robots.
This has motivated researchers in autonomous navigation and human-robot interaction to study different aspects of the navigation problem, such as navigation within crowds~\cite{chen2019crowd}, 
communicating implicit and explicit navigation information with haptic devices~\cite{che2020implicitexplicitdorsaokamura}, 
co-navigation~\cite{khambhaita2020viewingRachid} using anticipatory signals~\cite{unhelkar2015human},  navigation cues in human gaze~\cite{8990034_navCNNattentionhumangaze}, 
and the effects of distinct robot navigation strategies on pedestrians~\cite{mavrogiannis2019effectsnavstrategiesHRI}. 
Despite this long tradition, truly social mobile robots that are safe, efficient and fluent when deployed in human environments are still a research challenge.

Early solutions \textbf{\textbf{for navigation in human environments}} combine a motion planner~\cite{lavalle2006planning} with a controller that executes the plan~\cite{kunchev2006path}.
These approaches suffer from the ``freezing robot problem''~\cite{trautman2010unfreezing,trautman2015robot}, in which the planner finds all solutions to be below certain threshold of safety or feasibility, causing the robot to stop and continuously replan while the pedestrians move. This problem is more severe in domains with complex geometry and large number of pedestrians.
To overcome these limitations, researchers have resort to machine learning approaches~\cite{otte2015survey}. Multiple approaches have proposed to use imitation learning to match human navigation patterns~\cite{pomerleau1989alvinn}\cite{ muller2006off}\cite{bojarski2016end}\cite{pokle2019deep}. The limitation of imitation learning is that the agent does not generalize well beyond what has been seen in the demonstrations, resulting on potentially unsafe actions when the robot reaches states out of distribution. 
Reinforcement learning (RL)~\cite{sutton2018reinforcement} with a self-exploratory loop in simulation is a technique that naturally provides further coverage of the state space.

Previous works pioneered the use of \textit{\textbf{deep reinforcement learning} in navigation around pedestrians} and developed multi-agent collision avoidance systems focused on optimizing path efficiency (CADRL)\cite{chen2017decentralized}
 and using the reward function to induce social norms such as passing on the right (SA-CADRL)\cite{chen2017sociallyawareDRLmit_iros17}.
Deep RL has also been successfully used for crowd navigation with recursive architectures that model interactions among pedestrians in the crowd to improve the model's performance as the number of pedestrians becomes large in open spaces (SARL)\cite{chen2019crowd}. 
These approaches use pedestrians positions as input to the network (either ground-truth in simulation or detection/tracking from vision and lidar), 
but do not explicitly consider the presence of geometric constraints in the layout and are rather tested on spacious environments with no sensor input that reflects environmental constraints for the local interactions. This paper proposes a lidar-based framework 
with an architecture and training regime that enables the policy 
to acquire the behaviors needed 
to perform in constrained indoor environments.

Previous solutions have explored how to \textit{combine \textbf{motion planning and learning}} for navigation around dynamic obstacles. The idea is to leverage the motion planner to provide guidance to a learned controller, which adapts the plan to the current surrounding conditions. 
This multi-level strategy to combine the long-term planning (global path in the map) and short-term decision making also resembles the principles of how humans perform spatial navigation tasks \cite{epstein2017cognitive}.
\cite{pokle2019deep} and \cite{pmlr-v78-gao17a} showed this structure in imitation learning outperforms a single monolithic end-to-end architecture with no planning guidance, and \cite{sacnavicra2020} combined it with RL and active learning in order to learn human preferences for speed. We build on these ideas
and propose and architecture and training regime to use RL instead of imitation learning to avoid being limited to the regions of the state space covered in the expert demonstrations.

Finally, our learning procedure \textit{trains on simple instances} of a task with the goal of generalizing to task instances with the true complexity. Similar idea is behind methods on curriculum learning~\cite{Bengio2009CurriculumL} that create a series of increasingly complex task instances to facilitate learning~\cite{Matiisen2019TeacherStudentCL}\cite{graves2017automated}\cite{florensa2017reverse}. Differently, we assume that the challenges of our original navigation task cannot be learned progressively; they are the result of combining a finite set of complex patterns of layout-motion of dynamic agents (pedestrians). Therefore, we propose to learn to overcome these complex patterns separately in independent but intertwined training episodes, which will enables us to transfer eventually to a navigation task with the full combinatorial complexity.

\section{Learning to Navigate in Pedestrian Environments}
\label{sec:citations}

We propose to address the navigation problem in pedestrian environments with a combination of a sampling-base motion planner~\cite{lavalle2006planning} and a reactive low level sensorimotor policy learned with RL and implemented as a deep neural network. We assume that our solution can make use of a 2D map approximating the layout of static elements in the environment (e.g. furniture and walls) and that it can localize on it. Each episode the robot will start at a different location and is queried to navigate to a new final goal defined as a location in the map. We query the motion planner once at the beginning of the episode for a shortest path to the goal location. 
This path information is then used to define the low-level sensorimotor task that we describe next, followed by a description of the simulation environments we create to train our low-level policy.

\subsection{Reactive Navigation to Follow a Path}

We model the problem of navigating safely among humans following a given path as a Partially Observable Markov Decision Process (POMDP) defined by the tuple $\mathcal{M} = (\mathcal{S},\mathcal{A},\mathcal{O}, \mathcal{T},\mathcal{R},\gamma)$. Here, $\mathcal{S}$ is the state space; $\mathcal{A}$ is the action space; $\mathcal{O}$ is the observation space; $\mathcal{T}(s'|s,a), s\in\mathcal{S}, a\in\mathcal{A}$, is the state transition model defining a probability over the next state $s'$ after taking action $a$ in state $s$; $\mathcal{R}(s) \in \mathbb{R}$ is the reward at state $s$; $\gamma \in [0,1)$ is the discount factor. We assume that the state is not directly observable and learn a policy $\pi(a|o)$ conditioned on observations $o \in \mathcal{O}$ instead of latent states $s \in \mathcal{S}$.
The agent following the policy $\pi$ obtains an observation $o_t$ at time $t$ and performs an action $a_t$, receiving from the environment an immediate reward $r_t$ and a new observation $o_{t+1}$. Assuming the policy is parameterized by $\theta$, a policy gradient algorithm optimizes $\theta$ to maximize that the expected future return: 
\begin{equation}
\theta^* =  \underset{\theta}{\arg\max}\,J(\theta)= \underset{\theta}{\arg\max}\, E_\pi\left[\sum_{t}\gamma^t r_t\right]
\end{equation}

More concretely, the observation space, $\mathcal{O}$, includes the elements \mbox{$o = \{goal, lidar, waypoints\}$}, where $goal$ is the episodic navigation goal, represented by the 2D coordinates in robot's reference frame, $lidar$ contains the 128 range measurements from a 1D LiDAR sensor in robot sensor reference frame, and $waypoints$ contains a $n=6$ waypoints  computed by a global planner with access to a map of the environment. The action space, $\mathcal{A}$, defines actions \mbox{$a = \{v_x, v_y, \omega\}$}, where $(v_x, v_y)$ is the  commanded linear velocity, and $\omega$ is the commanded angular velocity for the mobile robot.

The reward function is composed of multiple terms that encourage the policy to reach the final goal, minimizing time, following the given global path but deviating from it to avoid the penalized collisions:
\begin{equation}
R = R_{goal} + R_{timestep} + R_{collision} + R_{potential} + R_{waypoint}
\end{equation}
where $R_{goal}=+1$ is a sparse reward assigned upon reaching the goal within a distance $d_g=0.5m$, $R_{timestep}=-0.001$ is a small negative reward with value per time step to encourage minimizing the task time, $R_{collision}=-1$ is assigned upon collision and terminates the episode, $R_{potential}$ is a dense reward assigned per time step as a function of the distance to the next desired waypoint.

\begin{figure}[t]
  \centering        
  \includegraphics[width=0.48\textwidth]{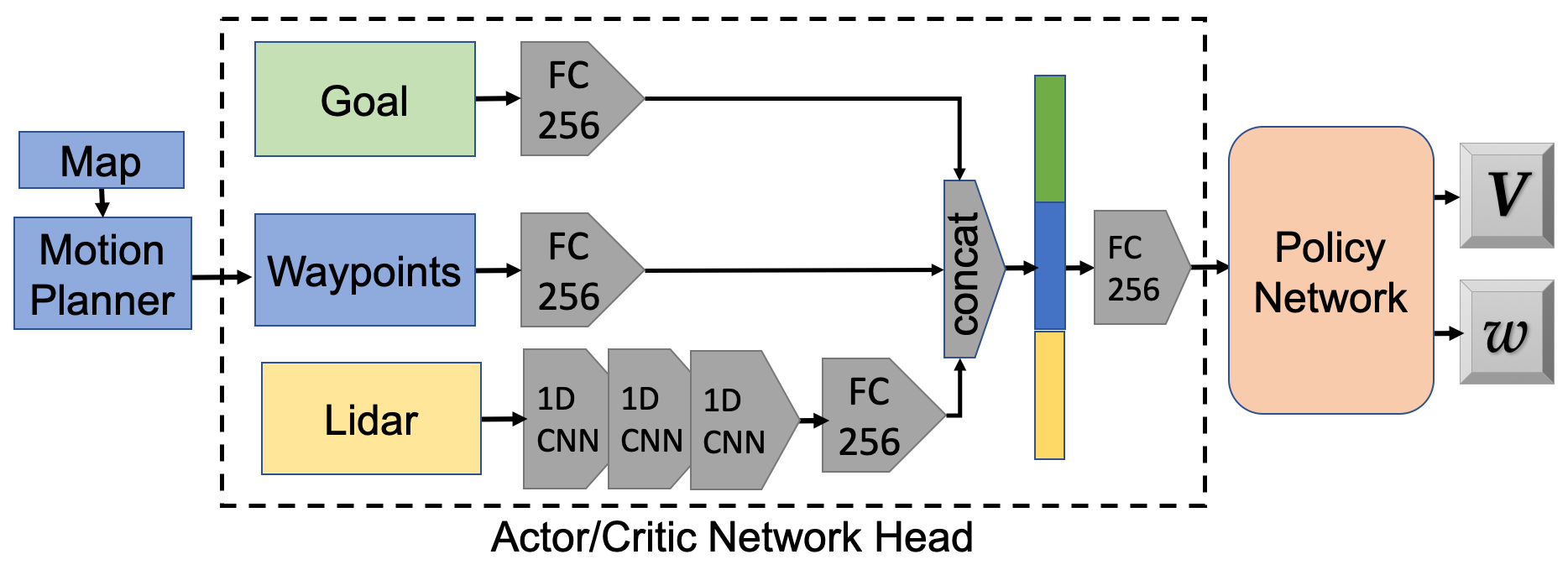}
  \caption[Placeholder]{Network Architecture. \vspace{-0.5cm}}
  \label{fig:model}
\end{figure}

We use the Soft Actor-Critic (SAC) algorithm~\cite{SAC_ICML_18,SAC_algapp_arxiv} to learn both a value function (critic) and a policy (actor) approximated by deep neural networks. SAC optimizes for a combined objective of the expected sum of rewards and the entropy.
The policy network maps from observation space to action space, providing velocity references to a low level velocity controller. The model architecture we propose for the policy network is depicted in Fig.\ref{fig:model}. The critic network presents a similar architecture with a different implementation after the shared featurizing head. Through extensive ablation studies, we tested a number of other variants for the observation space by including other inputs to the policy network, such as observation stacking with consecutive lidar frames, robot odometry, pedestrian detection (coordinates of pedestrians) and time to collision. These variants did not offer significant performance advantages in our domains and we opted for the simplest model that solves the task. However, these other inputs are expected to become relevant when considering explicit aspects of social interactions, such as pedestrian prediction, semantics, human-like social patterns, preferences and safety guaranties. We believe this is an interesting avenue for future research that can build on the presented model and leverage our multi-agent simulation environment.

\subsection{Simulated Environments}
\label{sec:sim}

We created a multi-agent simulation environment for the general problem of navigation around pedestrians, in which a simulated robot agent can collect experience and train according to the described POMDP. We use the Interactive Gibson Simulator (iGibson) \cite{48697interactivegibsonRAL} that runs on top of the pyBullet physics engine~\cite{coumans2016pybullet}. 
Figure \ref{fig:layoutsdefinition} shows a top-down view of the environments used for training and testing.  These layouts were designed to capture the essential geometric properties of many indoor environments such as corridors, crossing hallways and office spaces with doorways.  Many indoor spaces can be viewed as a composition of these simpler components.

Pedestrian behaviors are simulated using ORCA \cite{van2011reciprocalORCA}.
This model is often used to simulate pedestrian motion for crowd simulation and robot social navigation research \cite{chen2019crowd}.
ORCA uses a joint optimization and a centralized controller that guarantees that pedestrians will not collide with each other or any other objects added to a list of known obstacles, an useful property for driving the simulated pedestrians and provide a training environment for the robot. 

Based on a standard RL episodic training loop, a collision between the robot and any object results in termination of the episode and environment reset, which generates new random samples for the start and goal location of the robot and all pedestrians. Success is recorded if the robot gets to within 0.5 meters of the goal.  A timeout is recorded if it takes the robot longer than 125 seconds to get to the goal.  The personal space radius for each pedestrians was set to $10cm$.

\section{Experimental Evaluation and Results}
\label{sec:evaluation}

We conduct a series of experiments in simulation using the environments in Fig.\ref{fig:layoutsdefinition}
\footnote{{\small We have previously tested that policy networks that output linear and angular velocity commands can smoothly drive our robot platform and we expect successful real robot deployment in the future. In terms of sensing, the profile of the simulated lidar accurately reflect that of the real robot's.}}
\footnote{{\small We thank the Stanford HAI-AWS Cloud Credits for Research program for its support with compute resources.}}
. We design the experiments to study the following questions: 
\textbf{(Sec. \ref{sec:eval1})} Does the proposed approach enable a robot agent to learn to navigate around pedestrians in constrained environments?; and 
does multi-layout training on a small set of canonical environments (WALLS-ABCDF) generalize to navigation on more complex layouts (WALLS-GHI) that exhibit composition of the basic geometric elements? 
\textbf{(Sec. \ref{sec:eval2})} How does the proposed method compare with a planning-only approach; and with
\textbf{(Sec. \ref{sec:evalSARL})}
previous work on RL for a feasible domain? 
\textbf{(Sec. \ref{sec:eval3})} Can a policy trained in the simplified environments (WALLS-ABCDF) transfer to more complex 3D reconstructions of indoor environments (MESH)? 

The evaluation and comparison is conducted using a set of eight independently-trained policy networks. Two policies are trained using a \textbf{single layout} (\textbf{$\Pi_{WALLS-B}$}, \textbf{$\Pi_{WALLS-F}$}). Four policies are training using the proposed multi-layout training regime, in which the agent is randomly located in one of the layouts for each episode. The \textbf{multi-layout} polices are \textbf{$\Pi_{T1}$} trained on WALLS-ABF, \textbf{$\Pi_{T2}$} trained on WALLS-ABD, \textbf{$\Pi_{T3}$}  trained on  WALLS-ADF, and \textbf{$\Pi_{T4}$} trained on WALLS-ADE. Note that $\Pi_{T1}$ has been exposed to a corridor (A), door exit (B) and crosswalk (F), whereas $\Pi_{T2}$ doesn't see the crosswalk, $\Pi_{T3}$ doesn't see the door exit, and $\Pi_{T4}$ doesn't experience the crosswalk or door configurations during training. Finally, the MESH environments are used to train the policies \textbf{$\Pi_{MESH-MARKET}$} and \textbf{$\Pi_{MESH-HOME}$}, separately using the supermarket and apartment 3D reconstructions, respectively. 

For each experimental session, we train the policy for 200,000 training steps using 4 parallel simulation environments. We use the SAC implementation available on TFagents \cite{guadarrama2018tfagents}
and the available software integration with iGibson \cite{48697interactivegibsonRAL} and openAI Gym \cite{brockman2016openai}.

\textbf{Metrics:}
The following metrics are used to evaluate the objective performance of the system.
\textbf{Success rate (S)}: percentage  (\%) of episodes where the robot reaches its goal without colliding or timing out;
\textbf{Collision rate (C)}: percentage (\%) of episodes during which the robot collides with either a pedestrian or static object such as a wall;
\textbf{Collision with pedestrians (PC)}: percentage of episodes involving collisions with pedestrians;
\textbf{Collision with obstacles (OC)}: percentage (\%) of episodes involving collisions with static obstacles such as walls;
\textbf{Timeout rate (TO)}:  percentage (\%) of episodes where the robot runs out of time before reaching its goal;
\textbf{Personal Space Overlap (PSO)}: distance (cm) between the robot and pedestrians for which the robot is closest than the specified personal space threshold, aggregated across all pedestrians and averaged over length of the episode.

\subsection{Experiments}

\subsubsection{\textbf{Single- and Multi-Layout WALLS- Training}} \label{sec:eval1}
The single-layout and multi-layout policies are deployed in unseen layouts WALLS-G, WALLS-H and WALLS-I. Layout WALLS-H represents the generalization target with most compositional elements on it (corridors, doors, obstacles, intersections). The results are summarized in Tables \ref{table:singlelayout} and \ref{table:multilayout}, respectively. All presented Tables contain the average results over three experimental runs of 100 episodes each. Success rate below $90\%$ are highlighted in orange.

First, we note that policies from single-layout training, which achieve $S>90\%$ when tested in the same training environment, have a performance drop when tested on more complex layouts. Both $\Pi_{WALLS-B}$ and $\Pi_{WALLS-F}$ had lower performance in layout WALLS-H. This indicates the need for a larger body of experience during training, considering other arrangements of constraints and the patterns of pedestrian motion that arise in these configurations. Unlike single-layout training, using multiple canonical environments for training policy $\Pi_{T1}$ resulted in consistent generalization performance to all three test environments with average success rate of $95.16\%$ ($93.5\%$ WALLS-I, $96\%$ WALLS-H and $96\%$ WALLS-G) (Table \ref{table:multilayout}).

\begin{figure*}[t]
  \centering        
  \includegraphics[width=0.6\textwidth]{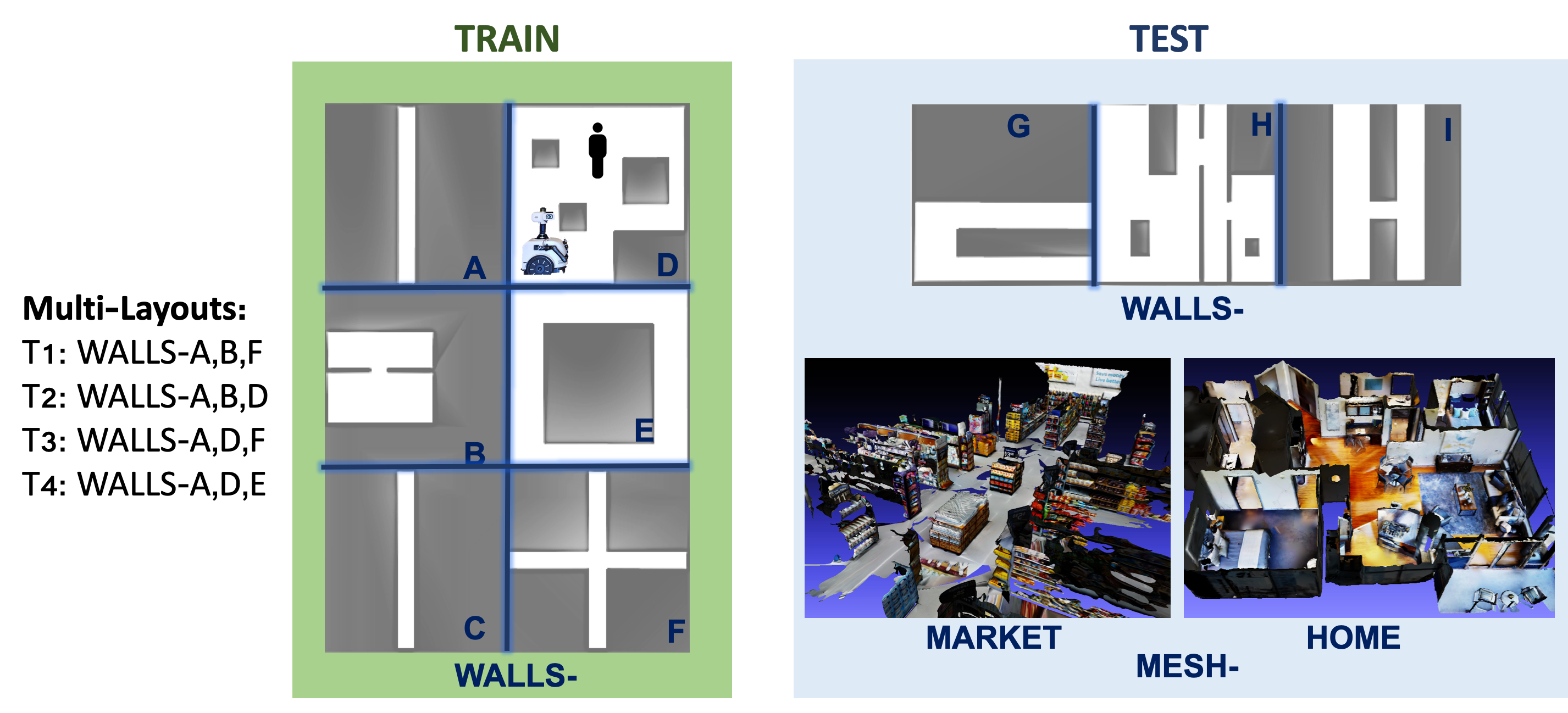}
  \caption{Definition of the layouts used in training and testing environments in simulation. \vspace{-3mm}}
  \label{fig:layoutsdefinition}
\end{figure*}

\input{table_singlelayouttrain}
\input{table_multilayout}

However, not all multi-layout policies generalize equally well to the compositional test environments. While $\Pi_{T1}$ and $\Pi_{T3}$ produce consistent transfer performance (on WALLS-G, WALLS-H, and WALLS-I), $\Pi_{T2}$ and $\Pi_{T4}$, which haven't experience the crosswalk or door configurations, have significant performance drop in WALLS-H which challenges the agent the most. $\Pi_{T4}$ also has a performance cost on WALLS-I. This results supports the importance of relevant geometric configurations in enabling transfer to a compositional domain, as a key element on learning beyond simply large collection of varied experience.

\begin{figure}[b]
  \centering        
  \includegraphics[width=0.5\textwidth]{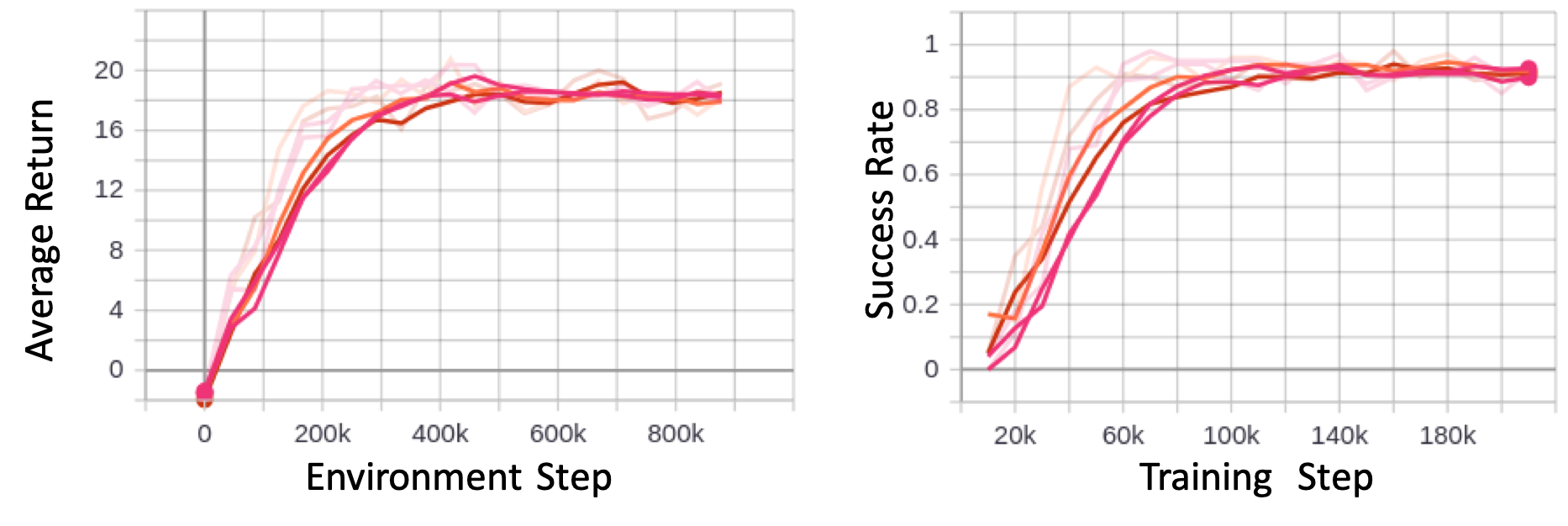}
  \caption{$\Pi_{T1}$ training curves: Average return as a function of environment steps and the success rate for interleaved evaluations every 10,000 training steps.}
  \label{fig:T1seeds}
\end{figure}

The multi-layout approach results in consistent training curves as shown in Figure \ref{fig:T1seeds} for  $\Pi_{T1}$ training with 4 random seeds. The resulting behavior of the policy can be observed in a video accompanying this paper  \href{https://ai.stanford.edu/~cdarpino/socialnavconstrained/}{linked here} \cite{PaperURLandVideo2020}. 

\subsubsection{\textbf{Comparison with a planning-based approach}} \label{sec:eval2}

\begin{figure}[t]
  \centering        
  \includegraphics[width=0.4\textwidth]{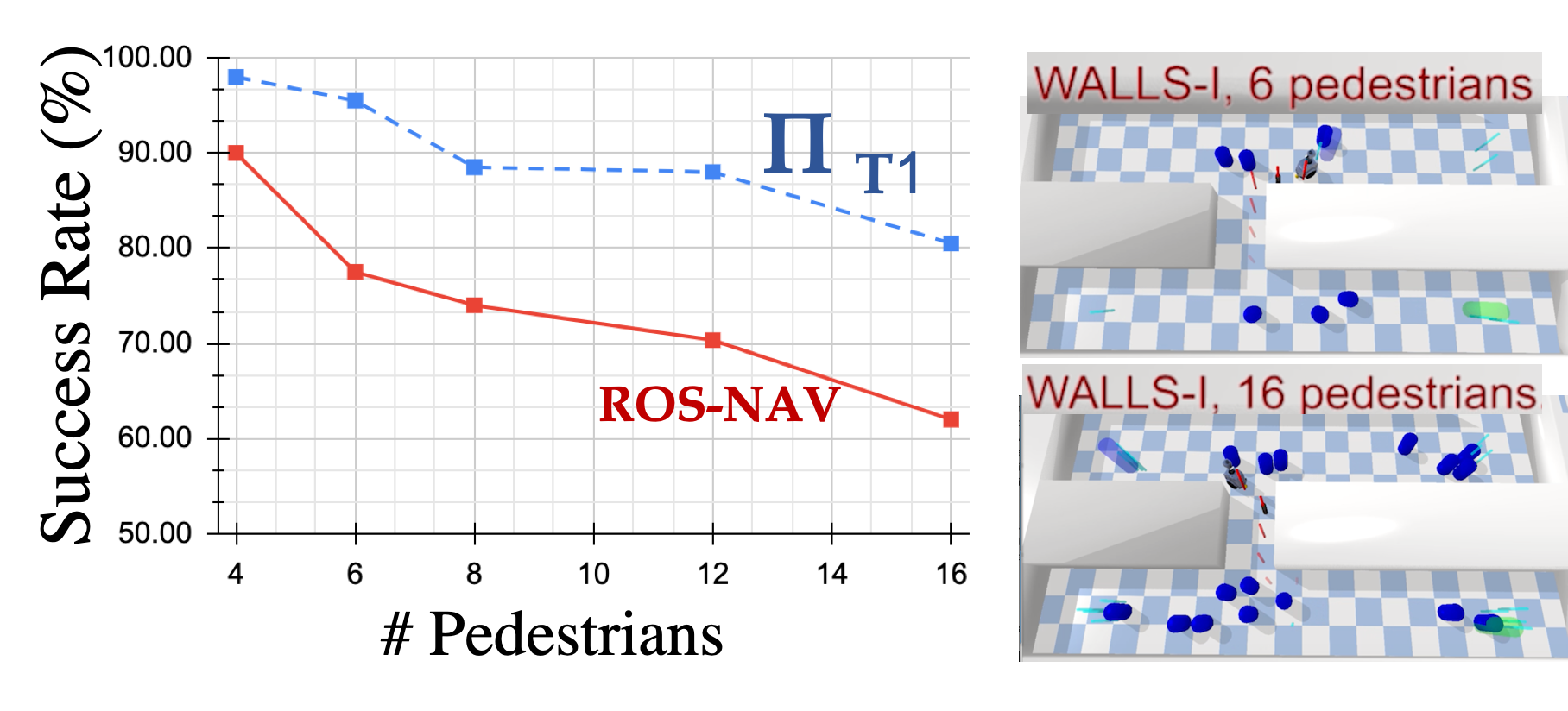}
  \caption{Performance of T1 and ROS-NAV tested on WALLS-I with increasing number of pedestrians. \vspace{-6mm}}
  \label{fig:pedswapplot}
\end{figure}

Planning approaches are suitable for application in our problem set up as they can also perform in arbitrary layouts (maps) without increasing complexity as a function of the layout geometry or size in an intractable manner.
We compare the performance of the learned policy $\Pi_{T1}$ with the planner available with the ROS Navigation framework (ROS-NAV-STACK)
. 
The planner in the ROS-NAV-STACK is a layered costmap method \cite{lu2014layered} that also uses two components: a global planner
using Dijkstra's algorithm and a local controller
that uses the dynamic window approach \cite{fox1997dynamicDWA}.
The ROS navigation parameters were tuned to optimize the performance of the planner in the WALLS layouts. The tests are conducted on WALLS-I with an increasing number of pedestrians in comparison with $\Pi_{T1}$. Results are summarized on Table \ref{table:rospedsawp} and the success rate as a function of the number of pedestrians is shown in Fig. \ref{fig:pedswapplot}. The planner resulted in lower success rate across all tests caused mainly due to timeouts. This robot-freezing problem \cite{trautman2010unfreezing} becomes more frequent as the pedestrian density increases and the planner fails to find feasible plans in the dynamic scene. While the  combination of planning and learning in $\Pi_{T1}$ has a performance that also degrades with the number of pedestrians, the failure cases are due to collisions and not timeouts. Note that real deployment can count with an additional sensing-based safety stop.

\input{table_meshtrain}
\input{table_ROSNAVbenchmark}

The comparison with a motion planner (no learning) is a proxy to quantify the benefit added by the RL components (Fig. \ref{fig:pedswapplot}). The planner is less successful at accomplishing the goal because it suffers frequently from the robot freezing problem  (large timeouts (TO) and low collisions -- due to not moving -- in Table \ref{table:rospedsawp}), which means that it continuously fails to find a feasible path and keeps re-planning. In contrast, out method resulted in no timeouts.

\subsubsection{\textbf{Comparison with a Reinforcement Learning Approach}} \label{sec:evalSARL}

\begin{figure}[t]
  \centering        
  \includegraphics[width=0.35\textwidth]{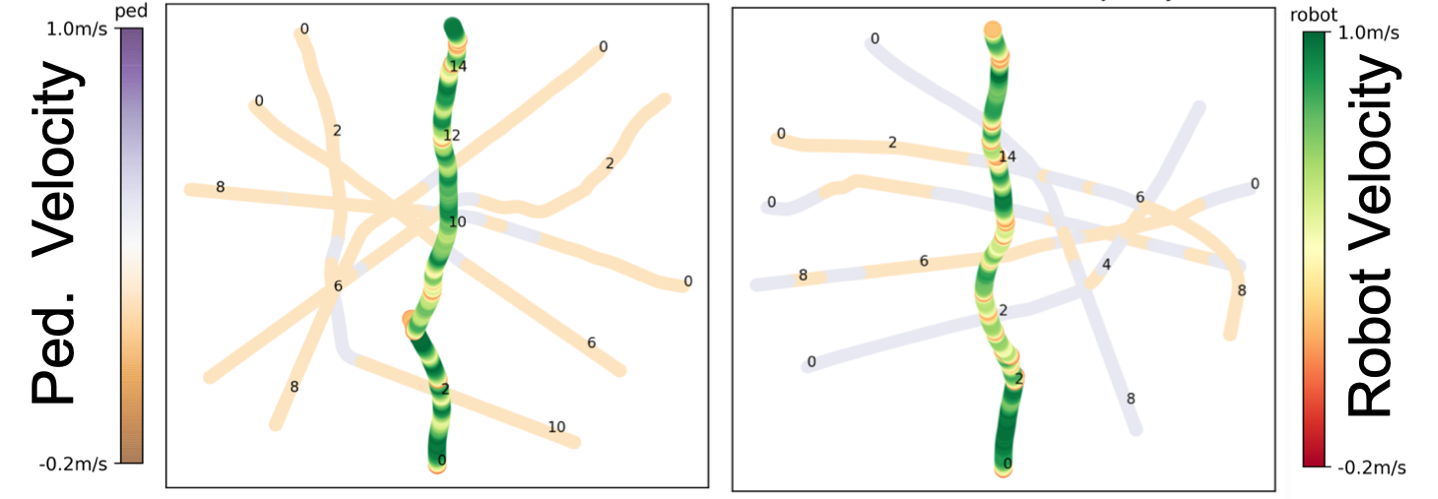}
  \caption{Example test episodes of $\Pi_{T1}$ in the circular open domain. \vspace{-5mm}}
  \label{fig:episodes}
\end{figure}

Since our target domain in constrained environments differs from previous works in open spaces, previous RL methods would require significant modifications to include the additional needed perceptual input to be comparable in our domain (refer to RL in Section \ref{sec:relatedwork}). To establish a baseline comparison with an existent RL architecture we 
test our policy $\Pi_{T1}$ on the domain
provided by \cite{chen2019crowd}, where SARL achieved state of the art results for crowd navigation
(including in comparison with relevant methods such as CADRL \cite{chen2017decentralized}).
SARL performs in open spaces and offers an open source implementation \cite{chen2019crowd}, including a widely-used comparative domain that consists of
a circle crossing scenario in which pedestrians starts at randomized positions on a circle with $4m$ radius and navigate to the antipodal point in the circle. The SARL policy is trained and tested on this domain \cite{chen2019crowd}, whereas we deploy our original policy $\Pi_{T1}$ which has not seen the circle domain. Both SARL and our policy achieve 100\% success rate in this domain, which establishes that our system is equally able to solve previous RL domains in the literature for the case of cooperative ORCA pedestrians. 
Analogous to SARL experiments in \cite{chen2019crowd}, Figure \ref{fig:episodes} shows some illustrative examples of $\Pi_{T1}$ navigation episodes. The robot lower speeds while pedestrians cross and then advances with increased speed. The policy allows linear velocities in the range $(-0.2,1)m/sec$, which enable stopping and backwards motions to accommodate pedestrians.

\subsubsection{\textbf{Cross-domain generalization with WALLS and MESH environments}} \label{sec:eval3}

We investigate if an agent trained in the simplified representation WALLS- is capable of navigating around pedestrians in realistic 3D scanned scenes (MESH), such as a supermarket (MESH-MARKET) and an apartment (MESH-HOME), illustrated in Fig. \ref{fig:layoutsdefinition} and shown in the accompanying video \cite{PaperURLandVideo2020}. 
First we measure the performance of training and testing on MESH-MARKET and MESH-HOME (i.e. MESH to MESH), and compare to that of $\Pi_{T1}$  in the mesh environments (i.e. multi-layout training on WALLS to MESH), and viceversa. The results are presented on Table \ref{table:mesh}.

The average success rate obtained by training and testing on the same mesh results in 79.5\% (MESH-MARKET =75\%; MESH-HOME=84\%), superior to the average transfer from WALLS to MESH of 71.5\% ($\Pi_{T1}$ in MESH-MARKET =77\%; $\Pi_{T1}$ in MESH-HOME=66\%). This is consistent with the expected performance of training in the target environment vs. policy transfer. However, transfer from MESH training to any other environments exhibits a significant performance drop compared to $\Pi_{T1}$. This indicates that the MESH performance is the product of overfitting to the environment, while WALLS training performs more consistently across domains. This result supports both the layout compositional argument and the approach of training on abstracted walls worlds \vspace{-2mm}.

\section{Conclusion}
\label{sec:conclusion}

We present an approach based on reinforcement learning to learn robot policies for navigation around pedestrians in constrained environments. The proposed model uses a motion planner that provides a globally planned trajectory, whereas the reinforcement component handles the local interactions needed for on-line adaptation to pedestrians. The learned policy exhibits interesting behaviors such as slowing down, maneuvering around pedestrians and going backwards to accommodate different common interactions that emerge as a consequence of the constrained layouts.

The analysis of the use of a compositionality principle for the proposed multi-layout training regime showed (1) the ability to train on simplified model abstractions and deploy on a more complex unseen environment with composition of the basic layouts; (2) that policies trained on relevant geometric configurations enable better generalization than those trained on layouts that don’t exhibit the geometry and inherent pedestrians interactions present in the target layout; and (3) policies trained on walls-worlds are able to generalize to unseen 3D reconstructions from real environments (such as an apartment and a supermarket). These results offer a practical advantage by removing the need to train on target environments with high reconstruction fidelity.




\include{supplementary_materials}

\end{document}

%% file: table_singlelayouttrain.tex
\begin{table*}[t]
\centering
\caption{Evaluation results using single-layout training \vspace{-2mm}}
\label{table:singlelayout}
\resizebox{0.7\textwidth}{!}{%
\begin{tabular}{llcrrrrrrrrrrr}
\cline{3-14}
 &
  \multicolumn{1}{l|}{} &
  \multicolumn{12}{c|}{\cellcolor[HTML]{EFEFEF}Train} \\ \cline{3-14} 
 &
  \multicolumn{1}{l|}{} &
  \multicolumn{6}{c|}{\cellcolor[HTML]{EFEFEF}\textbf{F}} &
  \multicolumn{6}{c|}{\cellcolor[HTML]{EFEFEF}\textbf{B}} \\ \hline
\multicolumn{1}{|r|}{\textbf{Test}} &
  \multicolumn{1}{c|}{\textbf{\# ped.}} &
  \multicolumn{1}{c|}{\textbf{S}} &
  \multicolumn{1}{c|}{\textbf{C}} &
  \multicolumn{1}{c|}{\textbf{PC}} &
  \multicolumn{1}{c|}{\textbf{OC}} &
  \multicolumn{1}{c|}{\textbf{TO}} &
  \multicolumn{1}{c|}{\textbf{PSO}} &
  \multicolumn{1}{c|}{\textbf{S}} &
  \multicolumn{1}{c|}{\textbf{C}} &
  \multicolumn{1}{c|}{\textbf{PC}} &
  \multicolumn{1}{c|}{\textbf{OC}} &
  \multicolumn{1}{c|}{\textbf{TO}} &
  \multicolumn{1}{c|}{\textbf{PSO}} \\ \hline
\multicolumn{1}{|r|}{\textbf{WALLS-I}} &
  \multicolumn{1}{c|}{\textbf{4}} &
  \multicolumn{1}{r|}{96.00} &
  \multicolumn{1}{r|}{4.00} &
  \multicolumn{1}{r|}{4.00} &
  \multicolumn{1}{r|}{0.00} &
  \multicolumn{1}{r|}{0.00} &
  \multicolumn{1}{r|}{2.07} &
  \multicolumn{1}{r|}{\cellcolor[HTML]{FFCE93}86.00} &
  \multicolumn{1}{r|}{14.00} &
  \multicolumn{1}{r|}{8.00} &
  \multicolumn{1}{r|}{6.00} &
  \multicolumn{1}{r|}{0.00} &
  \multicolumn{1}{r|}{1.96} \\ \hline
\multicolumn{1}{|r|}{\textbf{WALLS-H}} &
  \multicolumn{1}{c|}{\textbf{6}} &
  \multicolumn{1}{r|}{\cellcolor[HTML]{FFCE93}70.00} &
  \multicolumn{1}{r|}{30.00} &
  \multicolumn{1}{r|}{2.00} &
  \multicolumn{1}{r|}{28.00} &
  \multicolumn{1}{r|}{0.00} &
  \multicolumn{1}{r|}{0.66} &
  \multicolumn{1}{r|}{\cellcolor[HTML]{FFCE93}80.00} &
  \multicolumn{1}{r|}{20.00} &
  \multicolumn{1}{r|}{2.00} &
  \multicolumn{1}{r|}{18.00} &
  \multicolumn{1}{r|}{0.00} &
  \multicolumn{1}{r|}{1.30} \\ \hline
\multicolumn{1}{|r|}{\textbf{WALLS-G}} &
  \multicolumn{1}{c|}{\textbf{3}} &
  \multicolumn{1}{r|}{92.00} &
  \multicolumn{1}{r|}{8.00} &
  \multicolumn{1}{r|}{6.00} &
  \multicolumn{1}{r|}{2.00} &
  \multicolumn{1}{r|}{0.00} &
  \multicolumn{1}{r|}{1.09} &
  \multicolumn{1}{r|}{\cellcolor[HTML]{FFCE93}68.00} &
  \multicolumn{1}{r|}{32.00} &
  \multicolumn{1}{r|}{12.00} &
  \multicolumn{1}{r|}{20.00} &
  \multicolumn{1}{r|}{0.00} &
  \multicolumn{1}{r|}{2.80} \\ \hline
 &
   &
  \multicolumn{1}{l}{} &
  \multicolumn{1}{l}{} &
  \multicolumn{1}{l}{} &
  \multicolumn{1}{l}{} &
  \multicolumn{1}{l}{} &
  \multicolumn{1}{l}{} &
  \multicolumn{1}{l}{} &
  \multicolumn{1}{l}{} &
  \multicolumn{1}{l}{} &
  \multicolumn{1}{l}{} &
  \multicolumn{1}{l}{} &
  \multicolumn{1}{l}{} \\
 &
   &
  \multicolumn{1}{l}{} &
  \multicolumn{1}{l}{} &
  \multicolumn{1}{l}{} &
  \multicolumn{1}{l}{} &
  \multicolumn{1}{l}{} &
  \multicolumn{1}{l}{} &
  \multicolumn{1}{l}{} &
  \multicolumn{1}{l}{} &
  \multicolumn{1}{l}{} &
  \multicolumn{1}{l}{} &
  \multicolumn{1}{l}{} &
  \multicolumn{1}{l}{}
\end{tabular}%
}
\vspace{-8mm}
\end{table*}

%% file: table_multilayout.tex
\begin{table*}[th]
\centering
\caption{Evaluation results using multi-layout training \vspace{-2mm}}
\label{table:multilayout}
\resizebox{\textwidth}{!}{%
\begin{tabular}{rcrrrrrrrrrrrrllllll}
\cline{3-20}
\multicolumn{1}{l}{} &
  \multicolumn{1}{l|}{} &
  \multicolumn{6}{c|}{\textbf{$\Pi_{T1}$}} &
  \multicolumn{6}{c|}{\textbf{$\Pi_{T2}$}} &
  \multicolumn{6}{c|}{\textbf{$\Pi_{T3}$}} \\ \hline
\multicolumn{1}{|r|}{\textbf{Test}} &
  \multicolumn{1}{c|}{\textbf{\# ped.}} &
  \multicolumn{1}{c|}{\textbf{S}} &
  \multicolumn{1}{c|}{\textbf{C}} &
  \multicolumn{1}{c|}{\textbf{PC}} &
  \multicolumn{1}{c|}{\textbf{OC}} &
  \multicolumn{1}{c|}{\textbf{TO}} &
  \multicolumn{1}{c|}{\textbf{PSO}} &
  \multicolumn{1}{c|}{\textbf{S}} &
  \multicolumn{1}{c|}{\textbf{C}} &
  \multicolumn{1}{c|}{\textbf{PC}} &
  \multicolumn{1}{c|}{\textbf{OC}} &
  \multicolumn{1}{c|}{\textbf{TO}} &
  \multicolumn{1}{c|}{\textbf{PSO}} &
  \multicolumn{1}{c|}{\textbf{S}} &
  \multicolumn{1}{c|}{\textbf{C}} &
  \multicolumn{1}{c|}{\textbf{PC}} &
  \multicolumn{1}{c|}{\textbf{OC}} &
  \multicolumn{1}{c|}{\textbf{TO}} &
  \multicolumn{1}{c|}{\textbf{PSO}} \\ \hline
\multicolumn{1}{|r|}{\textbf{WALLS-I}} &
  \multicolumn{1}{c|}{\textbf{4}} &
  \multicolumn{1}{r|}
  {\cellcolor[HTML]{93ffce}93.50} &
  \multicolumn{1}{r|}{6.00} &
  \multicolumn{1}{r|}{3.50} &
  \multicolumn{1}{r|}{2.50} &
  \multicolumn{1}{r|}{0.50} &
  \multicolumn{1}{r|}{2.21} &
  \multicolumn{1}{r|}{92.00} &
  \multicolumn{1}{r|}{8.00} &
  \multicolumn{1}{r|}{2.00} &
  \multicolumn{1}{r|}{6.00} &
  \multicolumn{1}{r|}{0.00} &
  \multicolumn{1}{r|}{1.75} &
  \multicolumn{1}{r|}
  {\cellcolor[HTML]{93ffce}98.00} &
  \multicolumn{1}{r|}{2.00} &
  \multicolumn{1}{r|}{2.00} &
  \multicolumn{1}{r|}{0.00} &
  \multicolumn{1}{r|}{0.00} &
  \multicolumn{1}{r|}{1.99} \\ \hline
\multicolumn{1}{|r|}{\textbf{WALLS-H}} &
  \multicolumn{1}{c|}{\textbf{6}} &
  \multicolumn{1}{r|}
  {\cellcolor[HTML]{93ffce}96.00} &
  \multicolumn{1}{r|}{4.00} &
  \multicolumn{1}{r|}{3.33} &
  \multicolumn{1}{r|}{0.67} &
  \multicolumn{1}{r|}{0.00} &
  \multicolumn{1}{r|}{1.42} &
  \multicolumn{1}{r|}{\cellcolor[HTML]{FFCE93}88.00} &
  \multicolumn{1}{r|}{12.00} &
  \multicolumn{1}{r|}{6.00} &
  \multicolumn{1}{r|}{6.00} &
  \multicolumn{1}{r|}{0.00} &
  \multicolumn{1}{r|}{1.44} &
  \multicolumn{1}{r|}
  {\cellcolor[HTML]{93ffce}94.00} &
  \multicolumn{1}{r|}{6.00} &
  \multicolumn{1}{r|}{4.00} &
  \multicolumn{1}{r|}{2.00} &
  \multicolumn{1}{r|}{0.00} &
  \multicolumn{1}{r|}{1.51} \\ \hline
\multicolumn{1}{|r|}{\textbf{WALLS-G}} &
  \multicolumn{1}{c|}{\textbf{3}} &
  \multicolumn{1}{r|}
  {\cellcolor[HTML]{93ffce}96.00} &
  \multicolumn{1}{r|}{4.00} &
  \multicolumn{1}{r|}{4.00} &
  \multicolumn{1}{r|}{0.00} &
  \multicolumn{1}{r|}{0.00} &
  \multicolumn{1}{r|}{1.72} &
  \multicolumn{1}{r|}{92.00} &
  \multicolumn{1}{r|}{8.00} &
  \multicolumn{1}{r|}{4.00} &
  \multicolumn{1}{r|}{4.00} &
  \multicolumn{1}{r|}{0.00} &
  \multicolumn{1}{r|}{1.76} &
  \multicolumn{1}{r|}
  {\cellcolor[HTML]{93ffce}96.00} &
  \multicolumn{1}{r|}{4.00} &
  \multicolumn{1}{r|}{2.00} &
  \multicolumn{1}{r|}{2.00} &
  \multicolumn{1}{r|}{0.00} &
  \multicolumn{1}{r|}{1.85} \\ \hline
\multicolumn{1}{l}{} &
  \multicolumn{1}{l}{} &
  \multicolumn{1}{l}{} &
  \multicolumn{1}{l}{} &
  \multicolumn{1}{l}{} &
  \multicolumn{1}{l}{} &
  \multicolumn{1}{l}{} &
  \multicolumn{1}{l}{} &
  \multicolumn{1}{l}{} &
  \multicolumn{1}{l}{} &
  \multicolumn{1}{l}{} &
  \multicolumn{1}{l}{} &
  \multicolumn{1}{l}{} &
  \multicolumn{1}{l}{} &
  &
  &
  &
  &
  &
  \\ \cline{3-14}
\multicolumn{1}{l}{} &
  \multicolumn{1}{l|}{} &
  \multicolumn{6}{c|}{\textbf{$\Pi_{T4}$}} &
  \multicolumn{6}{c|}{\textbf{ROS-NAV}} &
  &
  &
  &
  &
  &
  \\ \cline{1-14}
\multicolumn{1}{|r|}{\textbf{Test}} &
  \multicolumn{1}{c|}{\textbf{\# ped.}} &
  \multicolumn{1}{c|}{\textbf{S}} &
  \multicolumn{1}{c|}{\textbf{C}} &
  \multicolumn{1}{c|}{\textbf{PC}} &
  \multicolumn{1}{c|}{\textbf{OC}} &
  \multicolumn{1}{c|}{\textbf{TO}} &
  \multicolumn{1}{c|}{\textbf{PSO}} &
  \multicolumn{1}{c|}{\textbf{S}} &
  \multicolumn{1}{c|}{\textbf{C}} &
  \multicolumn{1}{c|}{\textbf{PC}} &
  \multicolumn{1}{c|}{\textbf{OC}} &
  \multicolumn{1}{c|}{\textbf{TO}} &
  \multicolumn{1}{c|}{\textbf{PSO}} &
  &
  &
  &
  &
  &
  \\ \cline{1-14}
\multicolumn{1}{|r|}{\textbf{WALLS-I}} &
  \multicolumn{1}{c|}{\textbf{4}} &
  \multicolumn{1}{r|}{\cellcolor[HTML]{FFCE93}74.00} &
  \multicolumn{1}{r|}{26.00} &
  \multicolumn{1}{r|}{10.00} &
  \multicolumn{1}{r|}{16.00} &
  \multicolumn{1}{r|}{0.00} &
  \multicolumn{1}{r|}{1.98} &
  \multicolumn{1}{r|}{91.00} &
  \multicolumn{1}{r|}{3.00} &
  \multicolumn{1}{r|}{2.00} &
  \multicolumn{1}{r|}{1.00} &
  \multicolumn{1}{r|}{6.00} &
  \multicolumn{1}{r|}{0.00} &
  &
  &
  &
  &
  &
  \\ \cline{1-14}
\multicolumn{1}{|r|}{\textbf{WALLS-H}} &
  \multicolumn{1}{c|}{\textbf{6}} &
  \multicolumn{1}{r|}{\cellcolor[HTML]{FFCE93}78.00} &
  \multicolumn{1}{r|}{22.00} &
  \multicolumn{1}{r|}{6.00} &
  \multicolumn{1}{r|}{16.00} &
  \multicolumn{1}{r|}{0.00} &
  \multicolumn{1}{r|}{1.46} &
  \multicolumn{1}{r|}{\cellcolor[HTML]{FFCE93}89.00} &
  \multicolumn{1}{r|}{4.00} &
  \multicolumn{1}{r|}{0.00} &
  \multicolumn{1}{r|}{4.00} &
  \multicolumn{1}{r|}{7.00} &
  \multicolumn{1}{r|}{13.17} &
  &
  &
  &
  &
  &
  \\ \cline{1-14}
\multicolumn{1}{|r|}{\textbf{WALLS-G}} &
  \multicolumn{1}{c|}{\textbf{3}} &
  \multicolumn{1}{r|}{96.00} &
  \multicolumn{1}{r|}{4.00} &
  \multicolumn{1}{r|}{4.00} &
  \multicolumn{1}{r|}{0.00} &
  \multicolumn{1}{r|}{0.00} &
  \multicolumn{1}{r|}{1.05} &
  \multicolumn{1}{r|}{92.00} &
  \multicolumn{1}{r|}{3.00} &
  \multicolumn{1}{r|}{3.00} &
  \multicolumn{1}{r|}{0.00} &
  \multicolumn{1}{r|}{5.00} &
  \multicolumn{1}{r|}{0.00} &
  &
  &
  &
  &
  &
  \\ \cline{1-14}
\end{tabular}
}
\vspace{-5mm}
\end{table*}

%% file: table_meshtrain.tex
\begin{table*}[t]
\centering
\caption{Cross-domain generalization with WALLS and MESH environments \vspace{-2mm}}
\label{table:mesh}
\resizebox{\textwidth}{!}{%
\begin{tabular}{rc|r|r|r|r|r|r|r|r|r|r|r|r|r|r|r|r|r|r|}
\cline{3-20}
\multicolumn{1}{l}{} &
  \multicolumn{1}{l|}{} &
  \multicolumn{6}{c|}{\textbf{$\Pi_{T1}$}} &
  \multicolumn{6}{c|}{\textbf{MESH-MARKET}} &
  \multicolumn{6}{c|}{\textbf{MESH-HOME}} \\ \hline
\multicolumn{1}{|r|}{\textbf{Test}} &
  \textbf{\# ped.} &
  \multicolumn{1}{c|}{\textbf{S}} &
  \multicolumn{1}{c|}{\textbf{C}} &
  \multicolumn{1}{c|}{\textbf{PC}} &
  \multicolumn{1}{c|}{\textbf{OC}} &
  \multicolumn{1}{c|}{\textbf{TO}} &
  \multicolumn{1}{c|}{\textbf{PSO}} &
  \multicolumn{1}{c|}{\textbf{S}} &
  \multicolumn{1}{c|}{\textbf{C}} &
  \multicolumn{1}{c|}{\textbf{PC}} &
  \multicolumn{1}{c|}{\textbf{OC}} &
  \multicolumn{1}{c|}{\textbf{TO}} &
  \multicolumn{1}{c|}{\textbf{PSO}} &
  \multicolumn{1}{c|}{\textbf{S}} &
  \multicolumn{1}{c|}{\textbf{C}} &
  \multicolumn{1}{c|}{\textbf{PC}} &
  \multicolumn{1}{c|}{\textbf{OC}} &
  \multicolumn{1}{c|}{\textbf{TO}} &
  \multicolumn{1}{c|}{\textbf{PSO}} \\ \hline
\multicolumn{1}{|r|}{\textbf{MESH-MARKET}} &
  \textbf{6} &
  \cellcolor[HTML]{FFCE93}77.00 &
  23.00 &
  15.50 &
  7.50 &
  0.00 &
  1.96 &
  \cellcolor[HTML]{FFCE93}75.33 &
  24.67 &
  21.33 &
  3.33 &
  0.00 &
  2.19 &
  \cellcolor[HTML]{FFCE93}78.00 &
  20.00 &
  20.00 &
  0.00 &
  0.00 &
  1.54 \\ \hline
\multicolumn{1}{|r|}{\textbf{MESH-HOME}} &
  \textbf{6} &
  \cellcolor[HTML]{FFCE93}66.00 &
  34.00 &
  10.00 &
  24.00 &
  0.00 &
  1.13 &
  \cellcolor[HTML]{FFCE93}43.33 &
  56.67 &
  18.00 &
  38.67 &
  0.00 &
  1.66 &
  \cellcolor[HTML]{FFCE93}84.00 &
  16.00 &
  14.00 &
  2.00 &
  0.00 &
  1.93 \\ \hline
\multicolumn{1}{|r|}{\textbf{WALLS-H}} &
  \textbf{6} &
  {\cellcolor[HTML]{93ffce}96.00} &
  4.00 &
  3.33 &
  0.67 &
  0.00 &
  1.42 &
  \cellcolor[HTML]{FFCE93}88.00 &
  11.00 &
  6.00 &
  5.00 &
  1.00 &
  1.34 &
  \cellcolor[HTML]{FFCE93}64.00 &
  36.00 &
  9.00 &
  27.00 &
  0.00 &
  0.89 \\ \hline
\end{tabular}%
}
\vspace{-3mm}
\end{table*}

%% file: table_ROSNAVbenchmark.tex
\begin{table*}[t]
\centering
\caption{Comparison with ROS-NAV planner \vspace{-2mm}}
\label{table:rospedsawp}
\resizebox{0.6\textwidth}{!}{%
\begin{tabular}{rc|r|r|r|r|r|r|r|r|r|r|r|r|}
\cline{3-14}
\multicolumn{1}{l}{} &
  \multicolumn{1}{l|}{} &
  \multicolumn{6}{c|}{\textbf{$\Pi_{T1}$}} &
  \multicolumn{6}{c|}{\textbf{ROS-NAV}} \\ \hline
\multicolumn{1}{|r|}{\textbf{Test}} &
  \textbf{\# ped.} &
  \multicolumn{1}{c|}{\textbf{S}} &
  \multicolumn{1}{c|}{\textbf{C}} &
  \multicolumn{1}{c|}{\textbf{PC}} &
  \multicolumn{1}{c|}{\textbf{OC}} &
  \multicolumn{1}{c|}{\textbf{TO}} &
  \multicolumn{1}{c|}{\textbf{PSO}} &
  \multicolumn{1}{c|}{\textbf{S}} &
  \multicolumn{1}{c|}{\textbf{C}} &
  \multicolumn{1}{c|}{\textbf{PC}} &
  \multicolumn{1}{c|}{\textbf{OC}} &
  \multicolumn{1}{c|}{\textbf{TO}} &
  \multicolumn{1}{c|}{\textbf{PSO}} \\ \hline
\multicolumn{1}{|r|}{\textbf{WALLS-I}} &
  \textbf{4} &
  98.00 &
  2.00 &
  2.00 &
  0.00 &
  0.00 &
  1.69 &
  90.00 &
  2.50 &
  2.00 &
  0.50 &
  7.00 &
  0.00 \\ \hline
\multicolumn{1}{|r|}{\textbf{WALLS-I}} &
  \textbf{6} &
  95.50 &
  4.50 &
  4.00 &
  0.50 &
  0.00 &
  2.84 &
  \cellcolor[HTML]{FFCE93}{\color[HTML]{000000} 77.50} &
  3.50 &
  1.00 &
  2.50 &
  19.00 &
  0.00 \\ \hline
\multicolumn{1}{|r|}{\textbf{WALLS-I}} &
  \textbf{8} &
  \cellcolor[HTML]{FFCE93}88.50 &
  11.50 &
  10.00 &
  3.50 &
  0.00 &
  3.99 &
  \cellcolor[HTML]{FFCE93}{\color[HTML]{000000} 74.00} &
  3.00 &
  1.00 &
  2.00 &
  23.00 &
  1.35 \\ \hline
\multicolumn{1}{|r|}{\textbf{WALLS-I}} &
  \textbf{12} &
  \cellcolor[HTML]{FFCE93}88.00 &
  12.00 &
  11.50 &
  0.50 &
  0.00 &
  5.38 &
  \cellcolor[HTML]{FFCE93}{\color[HTML]{000000} 70.33} &
  2.67 &
  1.33 &
  1.33 &
  25.67 &
  0.62 \\ \hline
\multicolumn{1}{|r|}{\textbf{WALLS-I}} &
  \textbf{16} &
  \cellcolor[HTML]{FFCE93}80.50 &
  19.50 &
  19.00 &
  0.50 &
  0.00 &
  6.54 &
  \cellcolor[HTML]{FFCE93}{\color[HTML]{000000} 62.00} &
  2.67 &
  1.00 &
  1.67 &
  35.00 &
  0.97 \\ \hline
\end{tabular}%
}
\vspace{-5mm}
\end{table*}

%% file: supplementary_materials.tex
\section*{Supplementary Materials}

\setcounter{section}{0}
\setcounter{subsection}{0}


\section{Simulation Environment Implementation}

\paragraph{\textbf{Simulation}}
The robot collects experience in the simulated environment while being tasked to navigate from a set of initial to target positions. A policy network is learned using the off-policy Soft Actor Critic reinforcement algorithm (SAC) in continuous action space of robot velocity. The policy network receives guidance from a motion planner that provides waypoints to follow a globally planned trajectory, whereas the reinforcement component handles the balance with the local interactions. The RL agent is rewarded with a balance between following the waypoints but also deviating to accommodate for the presence of pedestrians. This framework results in policies that can adapt to arbitrary environments by generating a navigation motion plan, and leaves the local reactive interaction with nearby pedestrians to be learned by reinforcement. Robot perception is based on a lidar sensor, which raw data resulted in enough information for obstacles and pedestrians avoidance in the tested environments.

The simulation environment is built based on iGibson and the PyBullet physics engine. The learning agent is a differential drive mobile robot. These non-holonomic constraints are included in the simulation. 
The policy network outputs the desired values of linear and angular velocity for the robot $(V,w)$, which are passed to a low level velocity controller in wheel speed. This software architecture in simulation follows the architecture on the real robot in terms of input commands.

\paragraph{\textbf{Robot Behavior}}
The policy network receives as input a set of observations \mbox{$o = \{goal, lidar, waypoints\}$}, where $goal$ is the episodic navigation goal (2D polar coordinates of the goal in robot frame), $lidar$ contains 1D LiDAR sensor measurements in robot frame, and $waypoints$ contains a number of \textit{reference waypoints}  computed by a global planner with access to a map of the environment. The waypoints are selected from the collision-free shortest path between the start and goal points of the episode. The \textit{reference waypoints} are updated as the robot makes progress along the desired path. A \textit{waypoint resolution} is used to define the distance between consecutive waypoints, and a \textit{waypoint tolerance} value is used to define the minimum proximity from the robot to a waypoint to receive the waypoint reward and update the next desired waypoint. Parameters for the robot behavior are summarized on Table \ref{tab:sim}.

\input{table_simparam}

\begin{figure}[h]
  \centering        
  \includegraphics[width=0.4\textwidth]{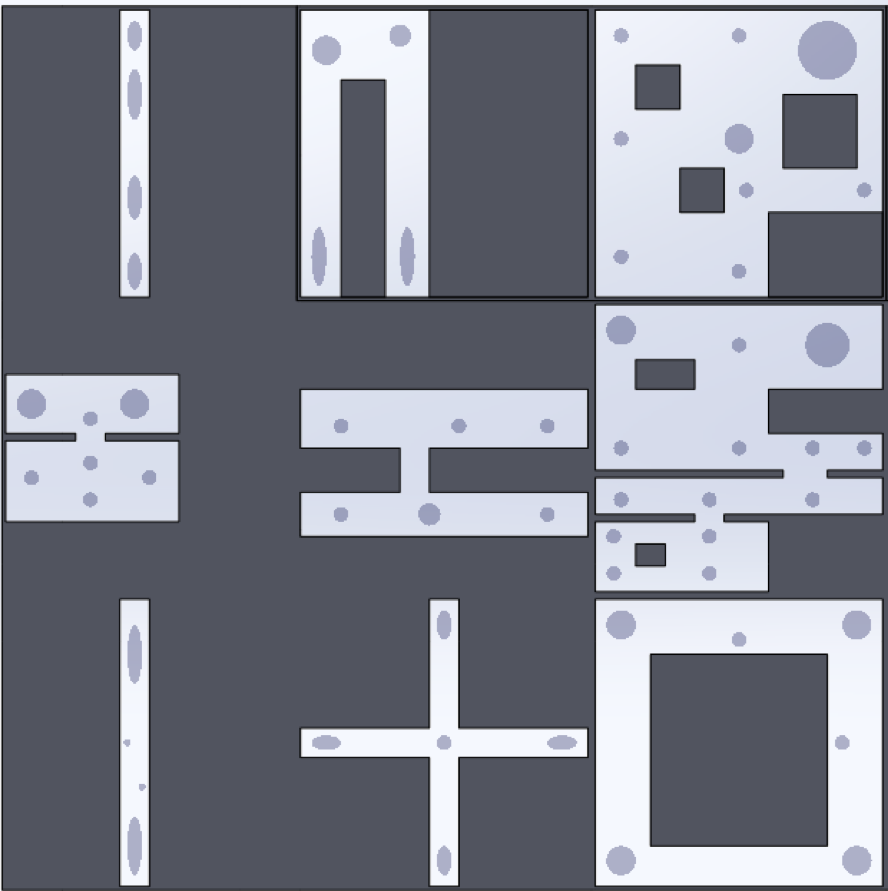}
  \caption{Locations on WALLS- domains. Each layout has dimensions $20m x 20m$.}
  \label{fig:wallslayoutsdetails}
\end{figure}

\paragraph{\textbf{Simulated Pedestrians}}
The simulated pedestrians follow the ORCA social forces model. The used parameters are summarized on Table \ref{tab:sim}. The ORCA model jointly optimizes the trajectories of all agents (pedestrians) to collaboratively avoid collisions. This property is guaranteed if all agents on the scene participate in the optimization. The ORCA model that drives all pedestrians has access to the state of the robot and considers it a participant agent. This produces the behaviors of the pedestrians to be collaborative with the robot under the expectation that the robot will be reciprocally collaborative in avoiding collisions. However, the robot follows a policy different from ORCA (either the learned policy or the ROS-NAV-STACK planner). This simulation strategy results in an approximation to mutual collaboration between the pedestrians and the robot. While ORCA offers an efficient model to simulate multiple pedestrians, it is limited in the types of trajectories the pedestrians can follow and does not adapt well to trajectory that require turns, concave obstacles or trajectories with lack of visibility to the goal. To address this limitation, each layout includes a definition of allowed initial and end areas (2D locations and  allowed radius shown in Fig. \ref{fig:wallslayoutsdetails} as shaded areas) to require underlying pedestrian trajectories that are feasible for the ORCA model to generate. For each episode, locations are randomly sampled from this allowed set. Pedestrians are represented in the simulator interchangeable as either a cylinder shape or a rigid mesh. We observe that improvements on the behavioral simulation of pedestrians at the macro level would potentially unlock greater adaptability of the learned behaviors to the real world. 
For example, in the simulation of MESH- worlds, which follows the same principles as the WALLS- environments, 
the simulation of pedestrians becomes more challenging as the specific context becomes relevant for sampling the start and end points for the ORCA-driven pedestrians: (e.g. walking from the kitchen to the bedroom).
We expect that improvement on pedestrian simulation alone will result on better robot performance overall,
without additional modifications to the method, as it would produce more fluent pedestrians trajectories. We leave these improvements as subject of future work.

\paragraph{\textbf{Multi-Layout Training}}
We train a set of policies on selected combinations of simple layouts WALLS-. These combinations are denominated $\{T1,T2,T3,T4\}$. For each $T_i$, the robot is randomly located in one of the included WALLS- layouts at the beginning of each training episode.

\paragraph{\textbf{Simulation Design}}
Using the proposed approach and generally on simulation-based robot learning, the engineering effort centers on designing a simulation environment that properly exhibits the dynamics and properties of the target domain. In robotics, this effort differs from planning techniques, in which the effort is rather centered in explicitly defining algorithmic solutions within some problem properties. Both approaches require significant design and engineering, while the difference lies in specifying the environment (the problem) or the algorithm (an explicit solution).

\section{Hyperparameters}

Table \ref{tab:sacparam} reports the hyper-parameters used for the reinforcement learning algorithm SAC. All the reported policies were learned using the same set of hyper-parameters.

We tuned the parameters of the ROS-NAV-STACK planner to optimize its performance in the tested environments. Adjusting these values reduces planner timeouts and increases likelihood of finding feasible solutions in the footprint of the tested layouts. Table \ref{tab:rosnavparamlist} reports the used parameters. 
\input{table_SAC}
\input{table_rosnavparam}

%% file: table_simparam.tex
\begin{table}[b]
\centering
\caption{Simulator, agent and pedestrians}
\label{tab:sim}
\begin{tabular}{lr}
\rowcolor[HTML]{EFEFEF} 
{\color[HTML]{000000} \textbf{Observations and Robot Behavior}} & \multicolumn{1}{l}{\cellcolor[HTML]{EFEFEF}{\color[HTML]{000000} }} \\
{\color[HTML]{000000} Angular Velocity}              & {\color[HTML]{000000} $[-0.5,0.5]rad/sec$}  \\
{\color[HTML]{000000} Linear Velocity}               & {\color[HTML]{000000} $[-0.2,1]m/sec$} \\
{\color[HTML]{000000} Number of Reference Waypoints} & {\color[HTML]{000000} 6}       \\
{\color[HTML]{000000} Waypoint Resolution}           & {\color[HTML]{000000} 1.00m}       \\
{\color[HTML]{000000} Waypoint Tolerance}            & {\color[HTML]{000000} 0.50m}       \\
\rowcolor[HTML]{EFEFEF} 
{\color[HTML]{000000} \textbf{Pedestrians Behavior (ORCA)}}     & \multicolumn{1}{l}{\cellcolor[HTML]{EFEFEF}{\color[HTML]{000000} }} \\
{\color[HTML]{000000} Personal Space}                & {\color[HTML]{000000} $10cm$}     \\
{\color[HTML]{000000} Pedestrian Radius}             & {\color[HTML]{000000} $30cm$}     \\
{\color[HTML]{000000} Prefered Velocity}             & {\color[HTML]{000000} $2m/sec$}   \\
{\color[HTML]{000000} Maximum Speed}                 & {\color[HTML]{000000} $2m/sec$}   \\
{\color[HTML]{000000} Robot Visible to Pedestrians}  & {\color[HTML]{000000} TRUE}      
\end{tabular}
\end{table}

%% file: table_SAC.tex
\begin{table}[h!]
\centering
\caption{SAC Hyperparameters}
\label{tab:sacparam}
\begin{tabular}{lr}
\rowcolor[HTML]{EFEFEF} 
{\color[HTML]{000000} \textbf{SAC Hyperparameters}} & {\color[HTML]{000000} }          \\
{\color[HTML]{000000} Hardware Configuration}        & {\color[HTML]{000000} NVIDIA® V100} \\
{\color[HTML]{000000} Optimizer}                    & {\color[HTML]{000000} Adam}      \\
{\color[HTML]{000000} Discount Ratio Gamma}         & {\color[HTML]{000000} 0.99}      \\
{\color[HTML]{000000} Critic Learning Rate}         & {\color[HTML]{000000} 0.0003}    \\
{\color[HTML]{000000} Actor Learning Rate}          & {\color[HTML]{000000} 0.0003}    \\
{\color[HTML]{000000} Alpha Learning Rate}          & {\color[HTML]{000000} 0.0003}    \\
{\color[HTML]{000000} Maximum Episode Lenght (Step)} & {\color[HTML]{000000} 500}          \\
{\color[HTML]{000000} Replay Buffer Capacity}       & {\color[HTML]{000000} 150000}    \\
{\color[HTML]{000000} Action Time Step}             & {\color[HTML]{000000} $0.25sec$}
\end{tabular}
\end{table}

%% file: table_rosnavparam.tex
\begin{table}[h!]
\centering
\caption{Optimized parameter for ROS-NAV planner}
\label{tab:rosnavparamlist}
\begin{tabular}{lr}
\rowcolor[HTML]{EFEFEF} 
{\color[HTML]{000000} \textbf{DWA Trajectory Scoring Parameters}} & {\color[HTML]{000000} }     \\
\rowcolor[HTML]{FFFFFF} 
{\color[HTML]{000000} path\_distance\_bias}           & {\color[HTML]{000000} 48}      \\
\rowcolor[HTML]{FFFFFF} 
{\color[HTML]{000000} goal\_distance\_bias}           & {\color[HTML]{000000} 24.00}   \\
\rowcolor[HTML]{FFFFFF} 
{\color[HTML]{000000} occdist\_scale}                 & {\color[HTML]{000000} 0.01}    \\
\rowcolor[HTML]{FFFFFF} 
{\color[HTML]{000000} }                               & {\color[HTML]{000000} }        \\
\rowcolor[HTML]{EFEFEF} 
{\color[HTML]{000000} \textbf{Global and local costmaps}}         & {\color[HTML]{000000} }     \\
\rowcolor[HTML]{FFFFFF} 
{\color[HTML]{000000} footprint\_padding}             & {\color[HTML]{000000} 0.05}    \\
\rowcolor[HTML]{FFFFFF} 
{\color[HTML]{000000} inflation\_layer: cost\_scaling\_factor}    & {\color[HTML]{000000} 2.58} \\
\rowcolor[HTML]{FFFFFF} 
{\color[HTML]{000000} inflation\_radius}              & {\color[HTML]{000000} 2.50}    \\
\rowcolor[HTML]{FFFFFF} 
{\color[HTML]{000000} }                               & {\color[HTML]{000000} }        \\
\rowcolor[HTML]{EFEFEF} 
{\color[HTML]{000000} \textbf{move\_base Parameters}} & {\color[HTML]{000000} }        \\
\rowcolor[HTML]{FFFFFF} 
{\color[HTML]{000000} planner\_frequency}             & {\color[HTML]{000000} 0.5 Hz}  \\
\rowcolor[HTML]{FFFFFF} 
{\color[HTML]{000000} controller\_frequency}          & {\color[HTML]{000000} 15 Hz}   \\
\rowcolor[HTML]{FFFFFF} 
{\color[HTML]{000000} planner\_patience}              & {\color[HTML]{000000} 200 sec} \\
\rowcolor[HTML]{FFFFFF} 
{\color[HTML]{000000} controller\_patience}           & {\color[HTML]{000000} 15 sec}  \\
\rowcolor[HTML]{FFFFFF} 
{\color[HTML]{000000} recovery\_behavior\_enabled}    & {\color[HTML]{000000} FALSE}   \\
\rowcolor[HTML]{FFFFFF} 
{\color[HTML]{000000} clearing\_rotation\_allowed}    & {\color[HTML]{000000} FALSE}   \\
\rowcolor[HTML]{FFFFFF} 
{\color[HTML]{000000} max\_planning\_retries}         & {\color[HTML]{000000} -1}     
\end{tabular}
\end{table}